\documentclass[final]{cvpr}
\usepackage{times}
\usepackage{epsfig}
\usepackage{graphicx}
\usepackage{gensymb}
\usepackage{amsmath,amssymb} 
\usepackage{graphicx}
\usepackage[utf8]{inputenc} 
\usepackage[T1]{fontenc}    
\usepackage{multirow}
\usepackage[font={small}]{caption}
\usepackage{comment}
\usepackage{amsfonts}       
\usepackage{nicefrac}       
\usepackage{microtype}      
\usepackage{url}            
\usepackage{booktabs}       
\usepackage{amsfonts}       
\usepackage{nicefrac}       

\usepackage{multirow}
\usepackage{xcolor}
\usepackage{subfig}
\usepackage[misc]{ifsym}

\usepackage[pagebackref=true,breaklinks=true,colorlinks,bookmarks=false]{hyperref}

\def\cvprPaperID{1845} 
\def\confYear{CVPR 2021}

\begin{document}

\title{Pose-Controllable Talking Face Generation by\\Implicitly Modularized Audio-Visual Representation}

\author{Hang Zhou$^{1}$, Yasheng Sun$^{2,3}$, Wayne Wu$^{2,4}$, Chen Change Loy$^{4}$, Xiaogang Wang$^{1}$, Ziwei Liu$^{4~\textrm{\Letter}}$\\
    $^1$CUHK - SenseTime Joint Lab, The Chinese University of Hong Kong \quad $^2$SenseTime Research\\
    $^3$Tokyo Institute of Technology  \quad
    $^4$S-Lab, Nanyang Technological University\\
      {\tt\small \{zhouhang@link,xgwang@ee\}.cuhk.edu.hk, wuwenyan@sensetime.com, \{ccloy,ziwei.liu\}@ntu.edu.sg }   \vspace{-20pt}
}

\twocolumn[{
\renewcommand\twocolumn[1][]{#1}%
\maketitle
\begin{center}
 \centering
 \includegraphics[width=0.95\textwidth]{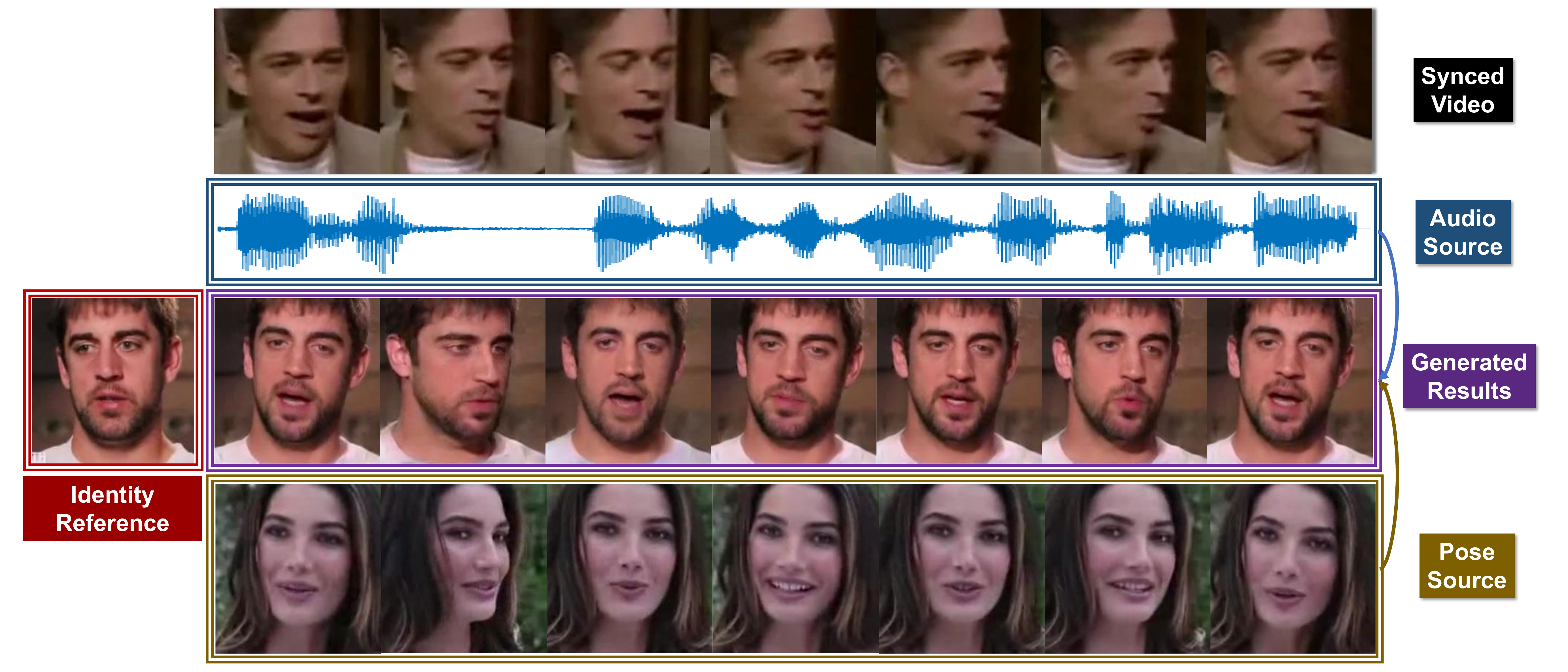}
\captionof{figure}{\textbf{Illustration of Pose-Controllable Audio-Visual System (PC-AVS).} Our approach takes one frame as identity reference and generates audio-driven talking faces with pose controlled by another \emph{pose source} video. The  mouth shapes of the generated frames are matched with the first row (synced video with audio) while the pose is matched with the bottom row (pose source).}
\label{fig:teaser}
\end{center}
}]

\maketitle


\begin{abstract}
While accurate lip synchronization has been achieved for arbitrary-subject audio-driven talking face generation, the problem of how to efficiently drive the head pose remains.
Previous methods rely on pre-estimated structural information such as landmarks and 3D parameters, aiming to generate personalized rhythmic movements. However, the inaccuracy of such estimated information under extreme conditions would lead to degradation problems. In this paper, we propose a clean yet effective framework to generate pose-controllable talking faces. We operate on non-aligned raw face images, using only a single photo as an identity reference. The key is to modularize audio-visual representations by devising an implicit low-dimension pose code. 
Substantially, both speech content and head pose information lie in a joint non-identity embedding space. While speech content information can be defined by learning the intrinsic synchronization between audio-visual modalities, we identify that a pose code will be complementarily learned in a modulated convolution-based reconstruction framework. 

Extensive experiments show that our method generates accurately lip-synced talking faces whose poses are controllable by other videos. Moreover, our model has multiple advanced capabilities including extreme view robustness and talking face frontalization.\footnote{Code, models, and demo videos are available at \url{https://hangz-nju-cuhk.github.io/projects/PC-AVS}.}
\end{abstract}

\section{Introduction}
\label{sec:intro}
Driving a static portrait with audio is of great importance to a variety of applications in the field of entertainment, such as digital human animation, visual dubbing in movies, and fast creation of short videos. 
%
Armed with deep learning, previous researchers take two different paths towards analyzing audio-driven talking human faces: 1) through pure latent feature learning and image reconstruction~\cite{chung2017you,zhu2018high,chen2018lip,zhou2019talking,song2018talking,vougioukas2019realistic,Prajwal2020lip}, and 2) to borrow the help of structural intermediate representations such as 2D landmarks~\cite{suwajanakorn2017synthesizing,chen2019hierarchical,das2020speech} or 3D representations~\cite{anderson2013expressive,thies2019neural,song2020everybody,chen2020talking,zhou2020makeittalk,richard2021audio,JiCVPR2021}.
Though great progress has been made in generating accurate mouth movements, 
most previous methods fail to model head pose, one of the key factors for talking faces to look natural.

It is very challenging to control head poses while generating lip-synced videos with audios. 
1) On the one hand, pose information can rarely be inferred from audios. While most previous works choose to keep the original pose in a video, very recently, a few works have addressed the problem of generating personalized rhythmic head movements from audios~\cite{chen2020talking,zhou2020makeittalk,yi2020audio}. 
However, they rely on a short clip of video to learn individual rhythms~\cite{yi2020audio,chen2020talking}, which might be inaccessible for a variety of scenes. 
2) On the other hand, all the above methods rely on 3D structural intermediate representations~\cite{yi2020audio,chen2020talking,zhou2020makeittalk}. 
%
%
The pose information is inherently coupled with facial movements, which affects both reconstruction-based~\cite{chung2017you,zhou2019talking} and 2D landmark-based methods~\cite{chen2019hierarchical,das2020speech}. 
Thus the most plausible way is to leverage 3D models~\cite{kim2018deep,thies2019neural,chen2020talking,Zhou_2020_CVPR} where the pose parameters and expression parameters are explicitly defined~\cite{blanz1999morphable}. Nevertheless, long-term videos are normally needed in order to learn person-specific renderers for 3D models. More importantly, such representations would be inaccurate under extreme cases such as large pose or low-light conditions.

In this work, we propose \textbf{Pose-Controllable Audio-Visual System (PC-AVS)}, which achieves free pose control when driving arbitrary talking faces with audios. Instead of learning pose motions from audios, we leverage another \emph{pose source} video to compensate only for head motions as illustrated in Fig.~\ref{fig:teaser}. Specifically, \emph{no structural information} is needed in our work. 
The key is to \emph{devise an implicit low-dimension pose code that is free of mouth shape or identity information}. In this way, audio-visual representations are \emph{modularized} into spaces of three key factors: speech content, head pose, and identity information.

In particular, we identify the existence of a non-identity latent embedding from the visual domain through data augmentation. 
Intuitively, the complementary \emph{speech content} and \emph{pose} information should 
originate from it. 
Extracting the shared information between visual and audio representations could lead to the \emph{speech content space} by synchronizing both the modalities,  which is also proven to be beneficial for various downstream audio-visual tasks~\cite{chung2016out,owens2018audio,zhou2019vision,Nagrani20d,gao2021visualvoice}.
However, there is no explicit way to model pose without precisely recognized structural information. 
Here we leverage the \emph{prior knowledge} of 3D pose parameters, that a mere vector of 12 dimensions, including a 3D rotation matrix, a 2D positional shifting bias, and a scale factor, is sufficient to represent a head pose. 
Thus we define a mapping from the non-identity space to a low dimension code which implicitly stands for the pose. 
Notably, we do not use other 3D priors to model the transformation between different poses. 
Then with additional identity supervision, the \emph{modularization} of the whole talking face representations has been completed.
%

The last key 
is the cross-frame reconstruction between video frames, where all representations are complementarily learned. 
A generator whose convolutional kernels are modulated by the embedded features is also designed. Specifically, we assemble the features from the \emph{modularized} spaces and use them to scale the weights of the kernels as proposed in~\cite{karras2020analyzing}.
The expressive ability of the weight modulation also enforces our \emph{modularization} to audio-visual representations in an implicit manner, \emph{i.e.},
in order to ensure low reconstruction loss, the low-dimensional code automatically controls pose while the speech content embedding takes care of the mouth.
%
During inference, we can drive an arbitrary face by a clip of audio with head movements controlled by another video. Multiple advanced properties can also be achieved such as extreme view robustness and frontalizing talking faces.

Our contributions are summarized as follows: \textbf{1)} We propose to modularize the representations of talking human faces into the spaces of speech content, head pose, and identity respectively, by devising a low-dimensional pose code inspired by 3D pose prior in talking faces. \textbf{2)}  The modularization is implicitly and complementarily learned in a construction-based framework with modulated convolution. \textbf{3)} Our model generates pose-controllable talking faces with accurate lip synchronization. \textbf{4)} As no structural intermediate information is used in our system, our model requires little pre-processing and is robust to input views.

\section{Related Work}

\noindent\textbf{Audio-Driven Talking Face Generation.}
Driving talking faces with audio input~\cite{chen2020comprises,zhu2020deep} has long been important research interest in both computer vision and graphics, where structural information and stitching techniques play a crucial role~\cite{bregler1997video,brand1999voice,wang2010synthesizing,wang2012high,zhou2018visemenet}. For example, Bregler \textit{et al.}~\cite{bregler1997video} rewrite the mouth contours. With the development of deep learning, different methods have been proposed to generate landmarks through time modeling~\cite{fan2015photo,fan2016deep,suwajanakorn2017synthesizing}. However, previous methods are mostly speaker-specific. 
Recently, with the development of end-to-end cross audio-visual generation~\cite{chen2017deep,zhao2018sound,gao20192,zhao2019sound,xu2019recursive,zhou2020sep,gan2020music,tian2021cyclic,xu2021audio,gan2020foley}, 
researchers have explored the speaker-independent setting~\cite{chung2017you,song2018talking,kr2019towards,zhou2019talking,zhou2020makeittalk}, which seeks a universal model that handles all identities with one or few frame references. 
After Chung \textit{et al.}~\cite{chung2017you} firstly propose an end-to-end reconstruction-based network, Zhou \textit{et al.}~\cite{zhou2019talking} further disentangle identity from word by adversarial representation learning. The key idea for reconstruction-based methods is to determine the synchronization between audio and videos~\cite{chung2016out,zhou2019talking,kr2019towards,Prajwal2020lip} where Prajwal \textit{et al.}~\cite{Prajwal2020lip} sync mouth with audio for inpainting-based reconstruction. However, these reconstruction-based works normally neglect head movements due to the difficulty in decoupling head-poses from facial movements.

As more compact and easy-to-learn targets, structural information is leveraged as intermediate representations within GAN-based reconstruction pipelines~\cite{chen2019hierarchical,das2020speech,thies2019neural,song2020everybody,zhou2020makeittalk,yi2020audio}. Chen \textit{et al.}~\cite{chen2019hierarchical} and Das \textit{et al.}~\cite{das2020speech} both use two-stage networks to predict 2D landmarks first and generate faces. But the pose and mouth are also entangled within 2D landmarks.
On the other hand, 3D tools~\cite{blanz1999morphable,deng2019accurate,bulat2017far,jiang2019disentangled} serve as strong intermediate representations. Zhou \textit{et al.} particularly model 3D talking landmarks with personalized movements, which generates the currently most natural results. 
However, free pose control is not achieved in their model. 
Chen~\textit{et al.} and Yi~\textit{et al.} both leverage 3D model to learn natural pose. However, their methods cannot render good quality under the ``one-shot'' condition. 
Moreover, the 3D fitting accuracy would drop significantly under extreme conditions. Here, we propose to achieve free pose control for one reference frame without relying on any intermediate structural information.

\noindent\textbf{Face Reenactment.}
Our work also relates to visually driven talking faces, which is studied in the realm of face reenactment. Similar to their audio-driven counterparts, most face reenactment works depend on structural information such as landmark~\cite{wayne2018reenactgan,huang2020learning,zhang2020freenet,zakharov2019few}, segmentation map~\cite{Chen_2020_CVPR,burkov2020neural} and 3D models~\cite{thies2016face2face,kim2018deep,kim2019neural}. Specifically, certain works~\cite{Wiles18a,siarohin2019first,wang2020one} learn the warping between pixels without defined structural information. 
While quite a number of papers~\cite{zakharov2019few,burkov2020neural,das2020speech} adopt a meta-learning procedure on a few frames for a new identity, our method does not require this procedure.

\section{Our Approach}

\begin{figure*}[t]
    \centering
    \includegraphics[width=0.97\linewidth]{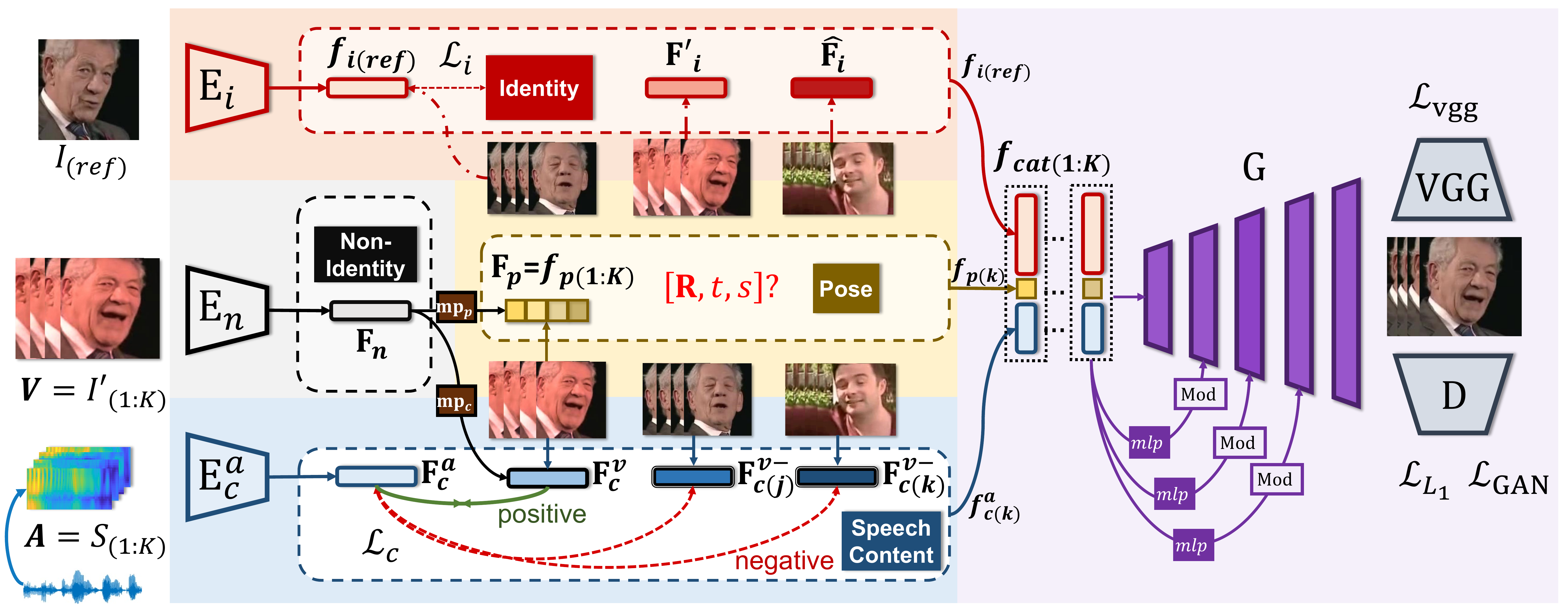}
    \vspace{-8pt}
    \caption{\textbf{The overall pipeline of our Pose-Controllable Audio-Visual System(PC-AVS) framework}. The identity reference $I_{(ref)}$ is encoded by $\text{E}_i$ to the \emph{identity space} (red).
    Encoder $\text{E}_n$ encodes video clip $\textbf{\textit{V}}$ to $\textbf{F}_n$ in the \emph{non-identity space} (grey). Then it is mapped to $\textbf{F}^v_c$ in the \emph{speech content space} (blue), which it shares with $\textbf{F}^a_c$ encoded by $\text{E}^a_c$ from audio spectrograms $\textbf{\textit{A}}$. We also draw encodings $\text{F}^{v-}_{c(j)}$, $\text{F}^{v-}_{c(k)}$ from two negative examples. 
    Their features in \emph{pose} and \emph{identity} spaces are also shown.
    Specifically, we map $\textbf{F}_n$ to pose features $\textbf{F}_p=f_{p(1:k)}$ in the \emph{pose space} (yellow). Though motivated by 3D priors, the pose features are not supervised by or necessarily represent the traditional $[\textbf{R}, t, s]$ 3D parameters. Finally, a pair of features $\{f_{i(i)}, f_{p(k)}, f^a_{c(k)}\}$ are assembled together and sent to generator $\text{G}$.
}
\label{fig:pipeline}
\vspace{-8pt}
\end{figure*}

We present \textbf{Pose-Controllable Audio-Visual System (PC-AVS)} that aims to achieve free pose control while driving static photos to speak with audio. The whole pipeline is depicted in Fig~\ref{fig:pipeline}. 
In this section, we first explore an efficient feature learning formulation by identifying the non-identity space (Sec.~\ref{sec:3.1}), then we provide the \emph{modularization} of audio-visual representations (Sec.~\ref{sec:3.2}). Finally, we introduce our generator and generating process (Sec.~\ref{sec:3.3}).

\begin{figure}[t]
    \centering
    \includegraphics[width=0.97\linewidth]{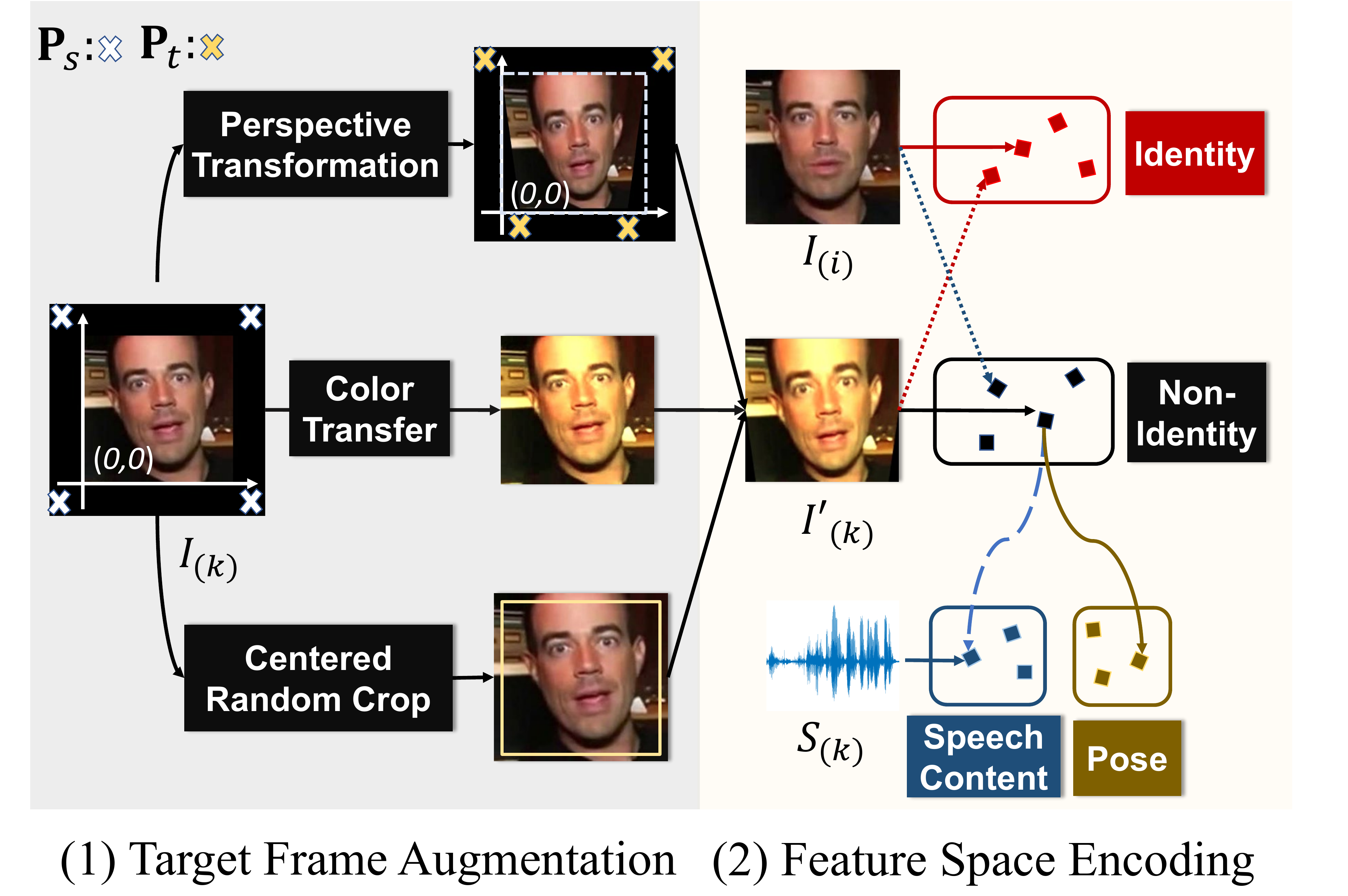}
    \vspace{-10pt}
    \caption{We \textbf{identify a non-identity space} through augmenting the (target) frames corresponding to the conditional audio. (1) Three data augmentation procedures are used to account for texture, facial deformation and subtle scale perturbation, which are irrelevant to learning pose and speech content. 
    (2) The feature spaces that we target at learning.
}
\vspace{-12pt}
\label{fig:augmentation}

\end{figure}

\subsection{Identifying Non-Identity Feature Space}
\label{sec:3.1}
At first, we revisit the general setting of previous pure reconstruction-based methods. The problem is formulated in an image-to-image translation manner~\cite{isola2017image,wang2018high}, where massively-available talking face videos provide natural self-supervision. 
%
Given  a $K$-frame video clip  $\textbf{\textit{V}} = \{I_{(1)}, \dots, I_{(K)}\}$, 
the natural training goal is to generate any \emph{target} frame $I_{(k)}$ conditioned on one frame of identity reference $I_{(ref)}$ ($ref \in [1, \dots, K]$) and the accompanied audio inputs.
%
The raw audios are processed into spectrograms  $\textbf{\textit{A}} = \{S_{(1)}, \dots, S_{(K)}\}$ as 2D time-frequency representations for more compact information preservation.
Previous studies~\cite{chen2018lip,zhu2018high,zhou2019talking,kr2019towards,Prajwal2020lip} have verified that learning the mutual and synchronized \emph{speech content} formation within both audio and visual modalities is effective for driving lips with audios. However, as no absolute pose information can be inferred from audios~\cite{Meishvili_2020_CVPR}, methods formulated in this way mostly keep the original pose unchanged.


In order to encode additional pose information, we first point out the existence of a general \emph{non-identity space} for representing all identity-repelling information including poses and facial movements. As depicted in Fig.~\ref{fig:augmentation}, the encoding of such a space is through careful data augmentation on the target frame $I_{(k)}$.
To account for two major aspects, 
namely texture and facial structure information, 
we apply two types of data augmentation to the target frames: \textit{color transfer} and \textit{perspective transformation}.
Additionally, a \textit{centered random crop} is also applied to alleviate the influence of facial scale changes in face detectors.  

The \textit{color transfer} is made by simply altering RGB channels randomly. As for the \textit{perspective transformation}, we set four source points
$
\mathbf{P}_s = [[-r_s, -r_s],~[\text{W} + r_s, -r_s],~[-r_s, \text{H} + r_s],~[\text{W} + r_s, \text{H} + r_s]]^{\text{T}}
$
outside of the original images by a random margin of $r_s$. Then through symmetrically moving the points along the $x$ axis, we set target points $
\mathbf{P}_t = [[-r_s + r_t, -r_s],~[ \text{W} + r_s, -r_s],~[-r_s, \text{H} + r_s],~[\text{W} + r_s + r_t, \text{H} + r_s]]^{\text{T}}
$ with another random step $r_t$~(see Fig.~\ref{fig:augmentation} (1) for details). The transformation can be learned by solving:
\begin{align}
    [\mathbf{P}_t, \mathbf{e}] = [\mathbf{P}_s, \mathbf{e}] * \mathbf{M},
\end{align}
where $\mathbf{M}$ is a $3 \times 3$ matrix, $*$ denotes the matrix multiplication and $\mathbf{e} = [1, 1, 1, 1]^{\text{T}}$.

In this \emph{non-identity space} lies the encoded features $\textbf{F}_n = \{f_{n(1)}, \dots, f_{n(K)}\}$ from the augmented target frames  $\textbf{\textit{V}}' = \{I'_{(1)}, \dots, I'_{(K)}\}$ by encoder $\text{E}_n$. 
Notably, data augmentation is also introduced in~\cite{burkov2020neural} for learning face reenactment. 
Different from their goal, 
our derivation of this space is to assist better feature learning and further representation modularization.


\subsection{Modularization of Representations}
\label{sec:3.2}
Supported with such a \emph{non-identity space}, we then modularize audio-visual information into three feature spaces namely the \emph{speech content space}, the \emph{head pose space} and \emph{identity space}.

\noindent\textbf{Learning Speech Content Space.} It has been verified that learning the natural synchronization between visual mouth movements and auditory utterances is valuable for driving images to speak~\cite{zhou2019talking,Prajwal2020lip}. 
Thus embedding space that contains synchronized audio-visual features as the \emph{speech content space}.

Specifically, we first define a mapping of fully connected layers 
from {non-identity} features  $\textbf{F}_n$ to the visual speech content features $\textbf{F}^v_c = \mathrm{mp}_c(\textbf{F}_n) = \{f^v_{c(1)}, \dots, f^v_{c(K)}\}$. Meanwhile, the audio inputs are encoded by the encoder $\text{E}^a_c$. Under our assumption, the audio features $\textbf{F}^a_c = \text{E}^a_c(\textbf{\textit{A}}) = \{ f^a_{c(1)}, \dots, f^a_{c(K)} \}$ share the \textbf{same space} with $\textbf{F}^v_c$. Thus the feature distance between timely aligned audio-visual pairs should be lower that non-aligned pairs.

We adopt the contrastive learning~\cite{chung2016out,Nagrani20d,zhou2019talking} protocol to seek the synchronization between audio and visual features. Different from the $L_2$ contrastive loss mostly leveraged before, we take the advantage of the more stable form of InfoNCE~\cite{oord2018representation}.
Concretely, for visual to audio synchronization, we regard the ensemble of timely aligned features $\textbf{F}^v_c \in \mathbb{R}^{l_c}$ and $\textbf{F}^a_c \in \mathbb{R}^{l_c}$ as positive pairs and sample $N^-$ negative audio features $\textbf{F}^{a-}_c \in \mathbb{R}^{N^- \times l_c}$. 
The negative audio clips could be sampled from other videos or from the same recording with a time-shift.
For feature distances measurement, we adopt the cosine distance $\mathcal{D}(\textbf{F}_1, \textbf{F}_2) = \frac{\textbf{F}_1^{\mathbf{T}} * \textbf{F}_2}{|\textbf{F}_1|\cdot|\textbf{F}_2|}$, where closer features render larger scores. In this way, the contrastive learning can be formulated to a classification problem with $(N^- + 1)$ classes:
\begin{align}
    \mathcal{L}^{v2a}_{c} = -\text{log}[\frac{\text{exp}({\mathcal{D}(\textbf{F}^v_c, \textbf{F}^a_c)})}{\text{exp}({\mathcal{D}(\textbf{F}^v_c, \textbf{F}^a_c)}) + \sum_{j=1}^{N^-}\text{exp}({\mathcal{D}(\textbf{F}^v_c, \textbf{F}^{a-}_{c(j)})})}].
\end{align}
The audio to visual synchronization loss $\mathcal{L}^{a2v}_{c}$ can also be achieved in a symmetric way as illustrated in the \emph{speech content space} of Fig~\ref{fig:pipeline}, which we omit here. The total loss for encoding this space is the sum of both:
\begin{align}
\mathcal{L}_{c} = \mathcal{L}^{v2a}_{c} + \mathcal{L}^{a2v}_{c}.
\end{align}

\noindent\textbf{Devising Pose Code.} 
Without relying on any precisely recognized structural information, such as pre-defined 3D parameters, it is difficult to explicitly model a pose. Here, we propose to  devise an implicit pose code using only subtle prior knowledge of 3D pose parameters~\cite{blanz1999morphable}. Concretely, the 3D head pose information can be expressed by a mere of 12 dimensions with a rotation matrix $\mathbf{R} \in \mathbb{R}^{3 \times 3}$, a positional translation vector $\mathbf{t} \in \mathbb{R}^2$ and a scale scalar $s$. Thus we define another fully connected mapping from the \emph{non-identity space} to a low-dimensional feature with exactly the size of 12: $\textbf{F}_p = \mathrm{mp}_p(\textbf{F}_n) = \{f_{p(1)}, \dots, f_{p(K)}\}$. 

The idea has some similarity with papers that use 3D priors for unsupervised 3D representation learning~\cite{HoloGAN2019,sitzmann2019scene}. Differently, we only use the prior knowledge on the minimum dimension of data needed. Though the implicitly learned feature could not possibly possess the same value of real 3D pose parameters, this intuition is important to our design. A pose code with larger dimensions may contain additional information that is not desired. The reason for the defined pose code to work is described in Sec.~\ref{sec:3.3}.

\noindent\textbf{Identity Space Encoding.} The learning of identity space 
has been well addressed in previous studies~\cite{chung2017you,zhou2019talking,zhou2020makeittalk}. As we train our networks on the videos of celebrities~\cite{Chung18b}, the identity labels naturally exist. Our identity space $ \textbf{F}_i = \text{E}_i(\textbf{\textit{V}}) = \{ f_{i(1)}, \dots, f_{i(K)} \}$ can be learned  on identity classification with softmax cross-entropy loss  $\mathcal{L}_i$ .

\subsection{Talking Face Generation}
\label{sec:3.3}
The features embedded in the three modularized spaces are composed for the final reconstruction of target frames $\textbf{\textit{V}}$. For a specific case, we concatenate $f_{i(ref)}$, $f^a_{c(k)}$ and $f_{p(k)}$ which are encoded from $I_{(ref)}$, $S_{(k)}$ and $I'_{(k)}$ respectively, and target to generate $I_{(k)}$ through a generator $\text{G}$.

\noindent\textbf{Generator Design.} A number of previous studies~\cite{chung2017you,zhou2019talking,kr2019towards,Prajwal2020lip} leverage skip connections for better input identity preserving. However, such a paradigm further restricts the possibility of altering poses, as the low-level information preserved through the connection greatly affects the generation results. Different from their structures, we directly encode the spatial dimension of the features to be one. With the recent development of generative model structures, style-based generator has achieved great success in the field of image generation~\cite{karras2019style,karras2020analyzing}. Their expressive ability in recovering details and style manipulation is also a crucial component of our framework.

In this paper, we empirically propose to generate faces through modulated convolution, which has been proven to be effective in image-to-image translation~\cite{park2020cut}. Detailedly, the concatenated features $f_{cat(k)} = \{f_{i(ref)}, f^a_{c(k)}, f_{p(k)}\}$ serve as latent codes to modulate the weights of the convolution kernels of the generator. At each convolutional block, a multi-layer perceptron is learned to map a $f_{cat(k)}$ to a  modulation vector $\mathcal{M}$ which has the same dimension as the input feature's channels. 
For each value $w_{xyz}$ in the convolution kernel weight $w$, where $x$ is its position on the input feature channels, $y$ is related to output channel numbers and $z$ represents the spatial location, it is modulated and normalized given the $x$'s value of $\mathcal{M}$ as: 
%
\begin{align}
    w^m_{xyz} = \frac{\mathcal{M}_x \cdot w_{xyz}}{\sqrt{{\sum}_{x,z} (\mathcal{M}_x \cdot w_{xyz})^2 + \epsilon}},
\end{align}
where $\epsilon$ is a small constant for avoiding numerical errors.


\setlength{\tabcolsep}{7pt}
\begin{table*}[t] 
\begin{center}  
\caption{\textbf{The quantitative results on LRW~\cite{chung2016lip} and VoxCeleb2~\cite{Chung18b}.} All methods are compared under the four metrics. For LMD the lower the better, and the higher the better for other metrics. $^{\dagger}$Note that we directly evaluate the authors' generated samples on VoxCeleb2 under their setting. They have not provided  examples on LRW.}
\vspace{-2pt}
\label{table:exp1}
\begin{tabular}{lcccccccc}
\toprule
 & \multicolumn{4}{c}{LRW~\cite{chung2016lip}}& \multicolumn{4}{c}{VoxCeleb2~\cite{Chung18b}} \\
\cmidrule(lr){2-5} \cmidrule(lr){6-9}
Method & $\text{SSIM}  \uparrow$&  $\text{CPBD}  \uparrow$  & LMD $\downarrow $ & $\text{Sync}_{conf}  \uparrow $ & $\text{SSIM}  \uparrow$&  $\text{CPBD}  \uparrow $  & LMD $ \downarrow$& $\text{Sync}_{conf}  \uparrow $\\

\midrule  
ATVG~\cite{chen2019hierarchical}  &0.810  & 0.102& 5.25  &4.1 &0.826 &0.061 &\textbf{6.49} &4.3 \\
Wav2Lip~\cite{Prajwal2020lip}   &\textbf{0.862} & 0.152&5.73 &\textbf{6.9} &0.846 &0.078 &12.26 &4.5 \\
MakeitTalk~\cite{zhou2020makeittalk}  &0.796  & 0.161& 7.13&3.1 &0.817 &0.068 &31.44 &2.8\\
Rhythmic Head$^{\dagger}$~\cite{chen2020talking} &-  & - & - &- &0.779 &0.802 &14.76 &3.8 \\
Ground Truth & 1.000 & 0.173 & 0.00 & 6.5 & 1.000 & 0.090 &0.00 &  5.9\\
\hline
Ours-Fix Pose & 0.815 & 0.180 &6.14 &6.3 &0.820 &\textbf{0.084} &7.68 & 5.8\\
\textbf{PC-AVS  (Ours)}  & \textbf{0.861} & \textbf{0.185} &\textbf{3.93} &6.4 &\textbf{0.886} &\textbf{0.083} & 6.88 &\textbf{5.9} \\
\bottomrule
\end{tabular}
\end{center}
\vspace{-12pt}
\end{table*}


\noindent\textbf{Network Training.} Finally, the feature space modularization and generator are trained jointly by image reconstruction. We directly borrow the same loss functions applied in~\cite{park2019semantic}. The generated and ground truth images are sent to a multi-scale discriminator $\text{D}$ with $N_D$ layers. 
The discriminator is utilized for both computing feature map $ L_1$ distances within its layers, and adversarial generative learning. The perceptual loss that relies on a pretrained VGG network with $N_P$ layers is also used. All loss functions can be briefly denoted as:
\begin{align}
\label{eq:4}
\begin{split}
    \mathcal{L}_{\text{GAN}} =& ~ \underset{\text{G}}{\text{min}}\underset{\text{D}}{\text{max}}\sum_{n=1}^{N_{D}}(\mathbb{E}_{I_{(k)}}[\log \text{D}_n(I_{(k)})]  \\ 
    & + \mathbb{E}_{f_{cat(k)}}[\log (1 - \text{D}_n(\text{G}(f_{cat(k)})]), \\
\end{split}\\
\label{eq:5}
    &\mathcal{L}_{L_1} =\sum_{i=1}^{N_{D}}{\|\text{D}_n(\text{I}_{(k)}) - \text{D}_{n}(\text{G}(f_{cat(k)})) \|_1 }, \\
\label{eq:6}
    & \mathcal{L}_{\text{vgg}} = \sum_{i=1}^{N_{P}}{\|\text{VGG}_n({I}_{(k)}) - \text{VGG}_{n}(\text{G}(f_{cat(k)})) \|_1 }.
\end{align}

This reconstruction training not only maps features to the image space but also implicitly ensures the representation modularization in a complementary manner. While features in the content space $\textbf{F}_c$ are synced audio-visual representations without pose information, in order to suppress the reconstruction loss, the low-dimension $f_p$ automatically compensates for pose information. 
Moreover, in most previous methods, the poses between generated images and their supervisions are not matched, which harms the learning process.
Differently in our setting, as the generated pose is aligned with ground truth through our pose code $f_p$, the learning of the speech content feature can further be benefited from the reconstruction loss. This leads to more accurate lip synchronization.

The overall learning objective for the whole system is formulated as follows:
\begin{align}
{\mathcal {L}}_{total} = {\mathcal{L}}_{\text{GAN}} + \lambda_1\mathcal{L}_{L_1} +  \lambda_v{\mathcal{L}}_{\text{vgg}} + \lambda_c\mathcal{L}_{c} + \lambda_i\mathcal{L}_{i},
\end{align}
where the $\lambda$s are balancing coefficients.

\section{Experiments}
\subsection{Experimental Settings}
\label{sec:4.1.1}
\noindent\textbf{Datasets.} We leverage two in-the-wild audio-visual datasets which are popularly used in a great number of previous studies, VoxCeleb2~\cite{Chung18b} and LRW~\cite{chung2016lip}. Both datasets provide detected and cropped faces. 
\begin{itemize}
    \item \textbf{VoxCeleb2}~\cite{Chung18b}. There are a total of 6,112 celebrities in VoxCeleb2 covering over 1 million utterances. While 5,994 speakers lie in the training set, 118 are in the test set. 
    The qualities of the videos differ largely. There are extreme cases with large head pose movements, low-light conditions, and different extents of blurry. 
    No test identity has been seen during training.
    \item \textbf{Lip Reading in the Wild (LRW)}~\cite{chung2016lip}. This dataset is originally proposed for lip reading. It contains over 1000 utterances of 500 different words within each 1-second video. Compared to VoxCeleb2, the videos in this dataset are mostly clean with high-quality and near-frontal faces of BBC news. Thus most of the utterances and identities in the test set are seen during training.
\end{itemize}

\label{sec:4.1.2}
\noindent\textbf{Implementation Details.} The structure of $\text{E}_i$ is a ResNeXt50~\cite{xie2017aggregated}. $\text{E}_n$ is borrowed from~\cite{zhou2019talking} and $\text{E}^a_c$ is a ResNetSE34 borrowed from~\cite{chung2020in}. The generator consists of 6 blocks of modulated convolutions. Different from certain previous works~\cite{chen2018lip,zhou2019talking,chen2020talking}, we do not align facial key points for each frame. All images are of size $224 \times 224$. The audios are pre-processed to 16kHz, then converted to mel-spectrograms with $\text{FFT}$ window size 1280, hop length 160 and 80 Mel filter-banks. For each frame, 0.2s mel-spectrogram with the target frame time-step in the middle are sampled as condition. The $\lambda$s are empirically set to 1.
Our models are implemented on PyTorch~\cite{paszke2019pytorch} with eight 16 GB Tesla V100 GPUs. %
Note that our identity encoder is pretrained on the labels provided in the Voxceleb2~\cite{Chung18b} dataset with $\mathcal{L}_i$. The speech content space is also pretrained first with loss $\mathcal{L}_c$. Then these models are loaded to the overall framework for generator and pose space learning. 

\begin{figure*}[t]
    \centering
    \includegraphics[width=0.98\linewidth]{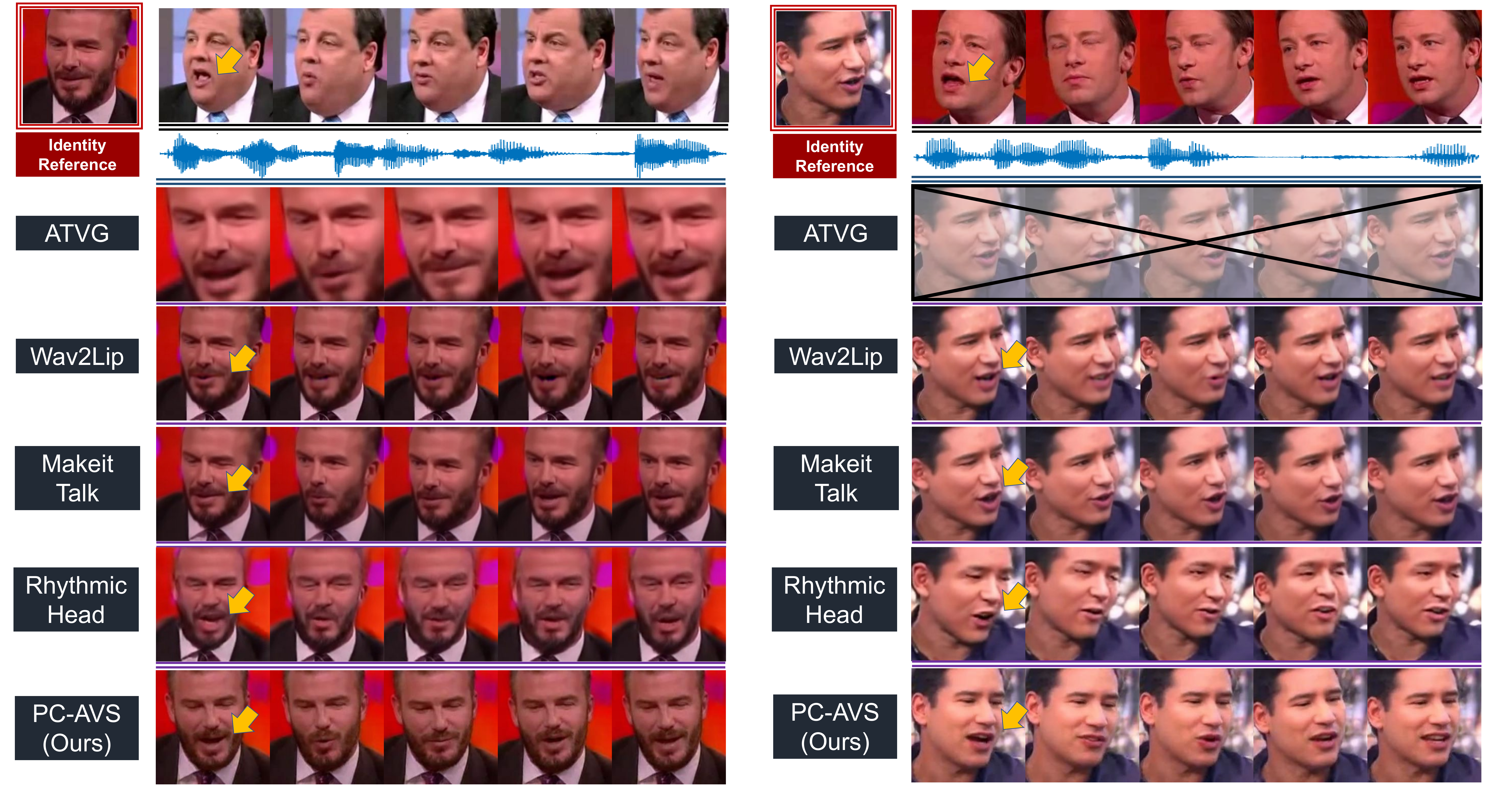}
    \vspace{-10pt}
    \caption{\textbf{Qualitative results}. In the top row are the audio-synced videos. ATVG~\cite{chen2019hierarchical} are accurate on the left. But with cropped faces, results seem non-real. Moreover, its detector fails on the right case. The mouth openings of Wav2Lip~\cite{Prajwal2020lip} are basically correct, but they generate static results. While MakeitTalk~\cite{zhou2020makeittalk} generates subtle head motions, the mouth shapes of their model are not accurate. Both our method and Rhythmic Head~\cite{chen2020talking} create diverse head motions, but their identify-preserving is much worse.
}
\label{fig:results}

\vspace{-5pt}
\end{figure*}

\label{sec:4.1.3}
\noindent\textbf{Comparison Methods.} We compare our methods with the best models currently available that support arbitrary-subject talking face generation. They are: \textbf{AVTG}~\cite{chen2019hierarchical}, the representative of 2D landmark-based method; 
\textbf{Wav2Lip}~\cite{Prajwal2020lip}, a reconstruction-based method that claims state-of-the-art lip sync results. 
\textbf{MakeitTalk}~\cite{zhou2020makeittalk} which leverages 3D landmarks and generates personalized head movements according to the driving audios. 
\textbf{Rhythmic Head}~\cite{chen2020talking} which generates rhythmic head motion under a different setting. Note that its
code is not run-able until this paper's final version, thus we only show two of its one-shot results in Fig.~\ref{fig:results}, which is generated with the help of the authors. The numerical comparisons in Table~\ref{table:exp1} are conducted on the authors' provided VoxCeleb2 samples under their setting for reference. Specifically, we also show the evaluation directly on the \textbf{Ground Truth}.

\setlength{\tabcolsep}{13pt}
\begin{table*}[t] 
\begin{center}

\caption{\textbf{User study measured by Mean Opinion Scores.} Larger is higher, with the maximum value to be 5.}
\vspace{-10pt}
\label{table:MOS}
\begin{tabular}{ccccc}

\hline

MOS on $\setminus$ Approach & ATVG~\cite{chen2019hierarchical}& Wav2Lip~\cite{Prajwal2020lip}  & MakeitTalk~\cite{zhou2019talking} & \textbf{PC-AVS (Ours)} \\
\noalign{\smallskip}
\hline

Lip Sync Quality & 2.87 & \textbf{3.98} & 2.33 & \textbf{3.98}\\
Head Movement Naturalness&1.26 &1.60&2.93& \textbf{4.20}\\

Video Realness& 1.60& 2.84&3.22 &\textbf{4.07}\\
\hline
\end{tabular}
\end{center}
\vspace{-12pt}
\end{table*}
\setlength{\tabcolsep}{1.4pt}

\subsection{Quantitative Evaluation}

\noindent\textbf{Evaluation Metrics.} We conduct quantitative evaluations on metrics that have previously been involved in the field of talking face generation. 
We use \textbf{SSIM}~\cite{wang2004image} to account for the generation quality, and 
the cumulative probability blur detection (\textbf{CPBD}) measure~\cite{narvekar2009no} is adopted from~\cite{vougioukas2019realistic} to evaluate the sharpness of the results.
Then we use both Landmarks Distance (\textbf{LMD}) around the mouths~\cite{chen2019hierarchical} and the confidence score ($\textbf{Sync}_{conf}$) proposed in SyncNet~\cite{chung2016lip}  to account for the accuracy of mouth shapes and lip sync. Though we do not use landmarks for video generation, we specifically detect landmarks for evaluation. 
Results on less informative metrics such as {PSNR} for image quality and {CSIM}~\cite{chen2020talking,zakharov2019few} for identity preserving are shown in supplementary material.

\noindent\textbf{Image Generation Paradigm.} We conduct the experiments under the self-driven setting that, the first image of each video within the test sets is selected as the identity reference. Then the audios are used as driving conditions to generate their accompanying whole videos. The evaluation is conducted between all generated frames and the ground truths. When the pose code is not given for our method, we can fix it with the same angle as the input, which keeps the head still. We refer results generated under this setting as \textbf{Ours-Fix Pose}. As our method is pose-controllable by another \emph{pose source} video,  we seek to leverage the pose information under a fair setting. 
Specifically, we use the target frames as pose sources to drive another reference image together with a different audio. In this way, an extra video with supposedly the same pose but different identities and mouth shapes is generated, serving as the pose source for our method.

\noindent\textbf{Evaluation Results.} The results are shown in Table~\ref{table:exp1}. It can be seen that our method reaches the best under most of the metrics on both datasets. 
On LRW, though Wav2Lip~\cite{Prajwal2020lip} outperforms our method given two metrics, the reason is that their method keeps most parts of the input unchanged while samples in LRW are mostly frontal faces.
Notably, their SyncNet confidence score outperforms that of ground truth's. But this only proves that their lip-sync results are nearly comparable to the ground truth.
Our model performs better than theirs on the LMD metric.
Moreover, the SyncNet confidence score of our results is also close to the ground truth on the more complicated VoxCeleb2 dataset, meaning that we can generate accurate lip-sync videos robustly.  

\subsection{Qualitative Evaluation}

Subject evaluation is crucial for determining the quality of the results\footnote{Please refer to \url{https://hangz-nju-cuhk.github.io/projects/PC-AVS} for demo videos and comparisons.} for generative tasks. Here we show the comparison of our methods against previous state-of-the-arts  (listed in Sec.~\ref{sec:4.1.3}) on two cases in Fig.~\ref{fig:results}. For our method, we sample one random \emph{pose source} video whose first frame's pose feature distance is closer to the input's. It can be seen that our method generates more diverse head motions and more accurate lip shapes. For the left case, only results of ATVG~\cite{chen2019hierarchical} and Rhythmic Head~\cite{chen2020talking} match our mouth shapes. Notably, the relied landmark detector~\cite{dlib09} of ATVG fails on the right case, thus no results can be produced. Similarly, the lip sync quality of MakeitTalk~\cite{zhou2020makeittalk} is better on the left side. The Face under the large pose on the right degrades the 3D landmarks' prediction. Both these facts verify the non-robustness of structural information-based methods. While producing dynamic head motions, the generated quality of Rhythmic Head's is much worse given only one identity reference.

\noindent\textbf{User Study.} We conduct a user study of 15 participants for their opinions on 30 videos generated by ours and three competing methods. Twenty videos are generated from fifteen reference images in the test set of VoxCeleb and ten from LRW.  The driving audios are also arbitrarily chosen from the test set in a cross-driving setting. Notably, we cannot generate Rhythmic Head~\cite{chen2020talking} cases without the help of the authors, this method is not involved. 
We adopt the widely used Mean Opinion Scores (MOS) rating protocol. The users are required to give their ratings (1-5) on the following three aspects for each video: (1) Lip sync qualities; (2) naturalness of head movements; and (3) the realness of results, whether they can tell the videos are fake or not.

The results are listed in Table~\ref{table:MOS}. As both ATVG~\cite{chen2019hierarchical} and Wav2Lip~\cite{Prajwal2020lip} generate near-stationary results, their scores on head movements and video realness are reasonably low. However, the lip sync score of Wav2Lip~\cite{Prajwal2020lip} is the same as ours, outperforming MakeitTalk~\cite{zhou2020makeittalk}. While the latter is famous for the realness of their generated image, the users prefer our results on the head motion naturalness and video realness, demonstrating the effectiveness of our method.

\setlength{\tabcolsep}{3pt}
\begin{table}[t] \footnotesize

\caption{\textbf{Ablation study with quantitative comparisons on VoxCeleb2~\cite{Chung18b}.} The results are shown when we vary the loss function. pose feature length and generator structure.}
\label{table:ablation}

\vspace{-10pt}
\begin{center}  
\begin{tabular}{lccccccc}
\toprule
Method & $\text{SSIM}  \uparrow$ & Mouth LMD $\downarrow $ & Pose LMD $\downarrow $ & $\text{Sync}_{conf}$$\uparrow $\\

\midrule  
w/o $\mathcal{L}_c$ &0.836 & 13.52 & 16.51 &4.7 \\
Pose-dim 36  &0.860 &9.17 &9.40 & 5.5  \\ 
AdaIN G  &0.750 &10.58& 8.78 & 5.5\\

\textbf{Ours}   &\textbf{0.886} & \textbf{6.88} & \textbf{5,9} &\textbf{7.62} \\
\bottomrule
\end{tabular}
\end{center}

\vspace{-15pt}
\end{table}
\begin{figure}[t]
    \centering
    \includegraphics[width=0.97\linewidth]{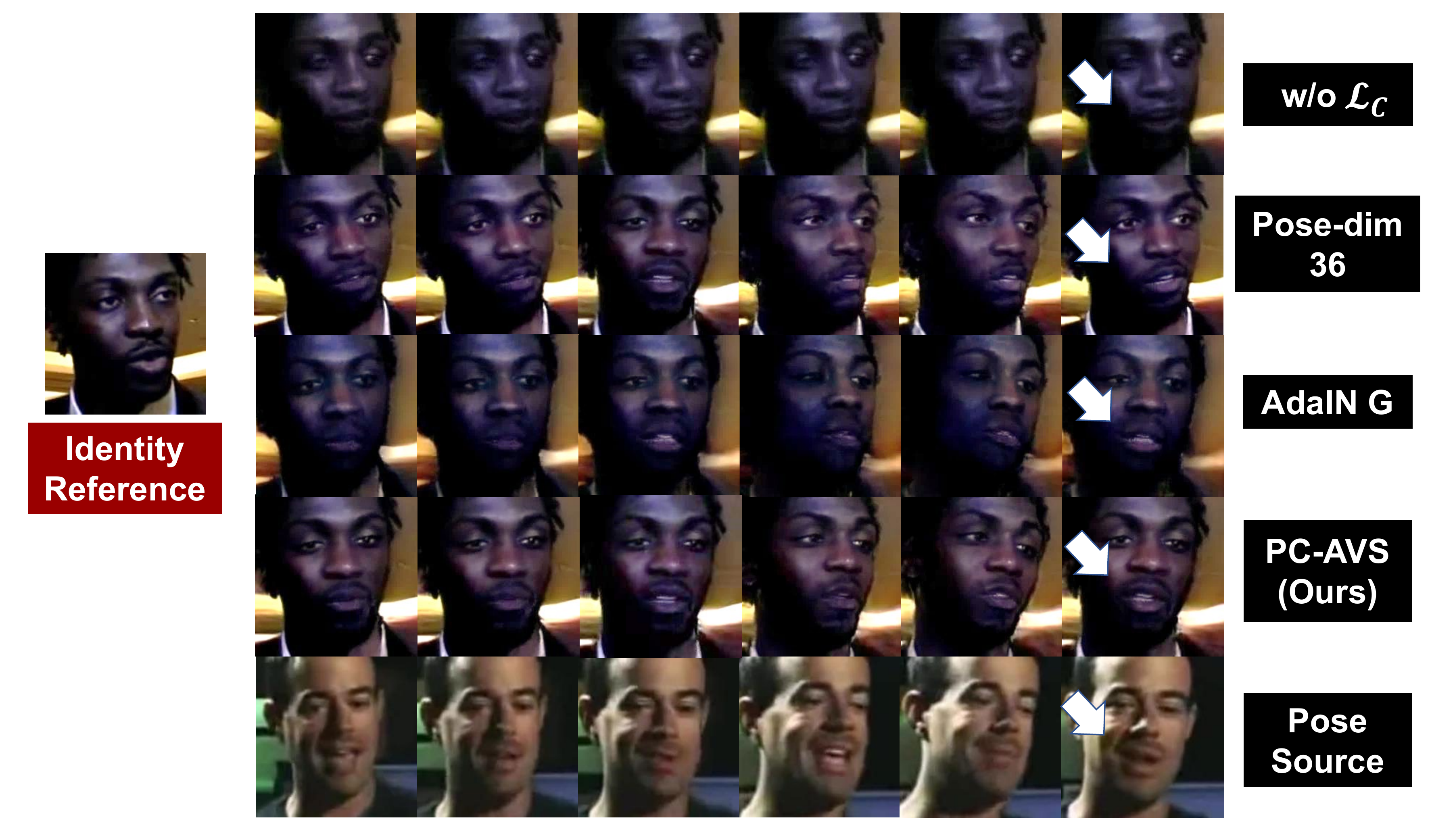}
    \vspace{-10pt}
    \caption{\textbf{Ablation study with visual results.} The mouth shapes are same among results but not synced with pose source.
}
\label{fig:ablation}

\vspace{-10pt}
\end{figure}

\subsection{Further Analysis} 
\noindent\textbf{Ablation Studies.} We conduct ablation studies given three important aspects of our method. The contrastive loss for audio-visual synchronization;The pose code length which is empirically set to 12 and design of the generator. 
Thus we conduct experiments on our model (1) w/o contrastive loss; (2) with different pose feature lengths and (3) change generator to AdaIN-based~\cite{karras2019style} form. Note that the traditional skip-connection form does not work in our case.
Except for the previous metrics, 
we propose an additional metric namely \emph{Pose LMD}, which computes only the landmark distances between facial contours to represent the pose.

The numerical results on VoxCeleb2 are shown in Table~\ref{table:ablation} and the visualizations are shown in Fig.~\ref{fig:ablation}.
Without the contrastive loss, the audios cannot be synced perfectly with the speech content, thus the whole modularization learning procedure would break, leading to the failure of pose code. 
Under the same training time, we observe drops on both metrics for larger pose code, possibly due to the growth of training difficulties along with the information in the pose code. Note that we also try learning without the target frame data augmentation (Sec.~\ref{sec:3.1}), similar to Ours w/o $\mathcal{L}_c$, the learning of the pose would fail.

\begin{figure}[t]
    \centering
    \includegraphics[width=0.97\linewidth]{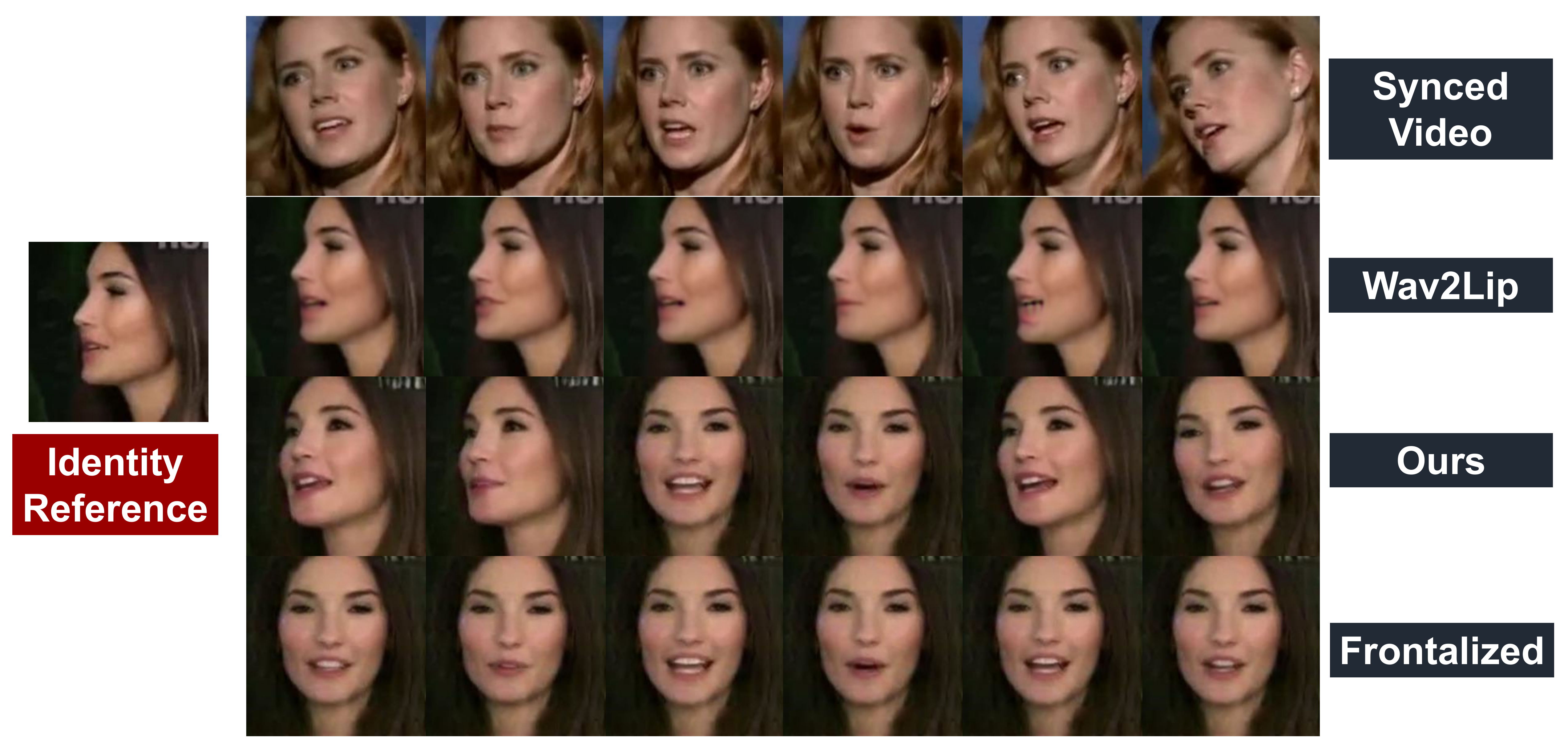}
    \vspace{-10pt}
    \caption{\textbf{Results under extreme condition.} We can drive faces under large poses, and even frontalize them by setting the pose code to all zeros.
}
\label{fig:additional}
\vspace{-15pt}
\end{figure}

\noindent\textbf{Extreme View Robustness and Face Frontalization.} Here we show the ability of our model to handle extreme views and achieving talking face frontalization. As ATVG fails again on the reference input of Fig.~\ref{fig:additional}, Wav2Lip would create artifacts on the mouth. Our model, on the other hand, not only can generate accurate lip motion but also can frontalize the faces while preserving the identity when setting the values of the pose code to zero.

\section{Conclusion}

In this paper, we propose \textbf{Pose-Controllable Audio-Visual System (PC-AVS)}, which generates accurately lip-synced talking faces with free pose control from other videos. We emphasize several appealing properties of our framework: 1) Without using any structural intermediate information, we implicitly devise a pose code and modularize audio-visual representations into the latent identity, speech content, and pose space. 
2) The complementary learning procedure ensures more accurate lip sync results than previous works. 3) The pose of our talking faces can be freely controlled by another pose source video, which can hardly be achieved before. 4) Our model shows great robustness under extreme conditions, such as large poses and viewpoints. \\

{\small
\noindent\textbf{Acknowledgements.} 
We would like to thank Lele Chen for his generous help with the comparisons, and Siwei Tang for his voice included in our video. 
Hang Zhou would also like to thank his grandmother, Shuming Wang, for her love throughout her life.
This research was conducted  in collaboration with SenseTime.
It is supported in part by the General Research Fund through the Research Grants Council
of Hong Kong under Grants (Nos. 14202217, 14203118, 14208619), in part by Research Impact Fund Grant No. R5001-18, and in part by NTU NAP and A*STAR through the Industry Alignment Fund - Industry Collaboration Projects Grant.}

{\small
\bibliographystyle{ieee_fullname}
\bibliography{egbib}
}

\end{document}